# An Automated News Bias Classifier Using *Caenorhabditis Elegans* Inspired Recursive Feedback Network Architecture

Agastya Sridharan[1], Natarajan S[2]
[1] 10th Grade,Scripps Ranch High School, San Diego, USA agastya.sridharan@gmail.com
[2]Professor, Department of Computer Science and Engineering, PES University, Bangalore, India natarajan@pes.edu

## Abstract

Traditional approaches to classify the political bias of news articles have failed to generate accurate, generalizable results. Existing networks premised on CNNs and DNNs lack a model to identify and extrapolate subtle indicators of bias like word choice, context, and presentation.

In this paper, we propose a network architecture that achieves human-level accuracy in assigning bias classifications to articles. The underlying model is based on a novel "Mesh" Neural Network (MNN) — this structure enables feedback and feedforward synaptic connections between any two neurons in the "mesh". The MNN contains six network configurations that utilize Bernoulli-based random sampling, pre-trained DNNs, and a network modelled after the C. Elegans nematode. The model is trained on over ten-thousand articles scraped from AllSides.com which are labelled to indicate political bias. The network's parameters are then evolved using a genetic algorithm suited to the feedback neural structure.

Finally, the best performing model is applied to five popular news sources in the United States over a fifty-day trial to quantify political biases in the articles they display. We hope our project can spur research into biological solutions for NLP tasks and provide accurate tools for citizens to understand subtle biases in the articles they consume.

## 1 Introduction

Prior to the rapid spread of social media and the Internet, there were a limited number of news sources that reported national events daily. Today, Americans consume thousands of news sites, many of which spin factual content to serve a political agenda. Studies by Robert Epstein et al. (2015) found that upon exposing over a thousand participants to one-sided news articles, subjects exposed to hyper-partisan content became significantly more distrustful of the political process. Moreover, this experience had "lasting effects" for years beyond the study. A similar experiment assessing the impact of polarized platforms concluded that biased articles made people "significantly more reticent to form consensus" on political issues (Dimock & Wike, 2020).

Therefore, it is imperative to develop an objective, automated method that can classify the political bias of news articles to inform readers of subtle biases in the articles they consume.

Various techniques have been employed to develop a bias classifier that utilize Bayesian algorithms, convolutional neural networks, and support vector machines, but none have produced results accurate enough to utilize in commercial applications. Part of the problem is the lack of generalizability; networks used for bias classification are incapable of recognizing subtle factors like context and word choice when exposed to unseen articles.

In this paper, we propose and implement three biologically-inspired models in order to minimize overfitting and generate generalizable, accurate outcomes: (1) a "mesh" neural network, (2) a genetic algorithm, and (3) a *C. Elegans* inspired synaptic configuration. We proceed to compare results from our model with state of the art networks and assess the network's capacity to generalize predictions. Lastly, we evaluate the structure and function of the genetic algorithm and discuss its viability for future NLP bias classification tasks.

## 2 Related Works

Related works for this paper encompass a broad spectrum covering technical papers, reviews and blogs.

There are two central problems with current automated approaches that seek to classify political bias in news articles: (1) discovering and identifying politically polarising text in a lengthy article, and (2) neutralising or reversing the polarising text without degrading its content or fluency. Liu et al. (2021), propose a unique paradigm to address these issues. Based on the Transformer architecture, they suggested a polarity flipper and a polarity detector. Human evaluations of their model demonstrate that the network is capable of reversing and neutralizing both headline and full-length articles while retaining overall fluency and substance. Tests indicate that their model functions effectively, but there is still significant potential for improvement.

Another approach has been to analyse the frequencies with which the media uses different phrases. Samantha D'Alonzo and Max Tegmark (2021) created a model based on the conditional probability distribution of phrases in nearly a million articles from around a hundred newspapers to analyse bias in dozens of news topics. The mapping of newspapers into a two-dimensional bias landscape based on their method agreed well with previous bias classifications based on human judgment.

Both networks produced accurate test set results but failed to account for outliers where subtle context clues altered the meaning of subsequent text. Their feature-based algorithms rely on the distribution of low-level lexical information, which often fail to identify context clues that alter the meaning of subsequent text. A report by Wei-Fan Chen et. al (2020) demonstrates that second-order information about bias statements helps to improve detection effectiveness when coupled with probability distributions that quantify the frequency, positions, and sequential order of lexical and informational sentence-level bias built in a Gaussian Mixture Model. Lechner et al. (2020) suggests this second-order information can be achieved through a combination of brain inspired neural computing as well as scalable deep learning architectures. This can create a single algorithm which caters both for generalizability by learning coherent representations as well as interpretable explanations of its dynamics.

One intriguing brain-inspired neural computing matrix that could satisfy Lechner’s guidelines is the Caenorhabditis elegans worm – the only organism whose neural capacitors have been fully mapped. *C. Elegans* has six touch receptors, five pairs of interneurons, and 69 motor neurons in its brain circuits for touch-induced movement. Johanne Hiandis et. al (2016) notice the coexistence of coherent and incoherent activity in the worm, known as chimera states, which have previously been studied in complex networks. However, they have not been investigated in modular networks. They look at a neural network that was modelled after the soil worm *C. Elegans*' connectome. The network is divided into six interconnected communities, and its neurons adhere to chaotic bursting dynamics. They hypothesize that electrical and chemical synapses connect neurons both inside and across their populations. According to numerical simulations, the interaction between these two types of coupling can change the dynamics in a way that allows for the emergence of chimera-like states (Deng et. al, 2016). They are made up of a mixture of desynchronized neurons that are a part of smaller communities and a mixture of synchronized neurons that are a part of the bigger communities. Mathias Lechner et. al (2020), report the development of a new liquid time-constant recurrent neural network for robotic control that is modelled after the nematode’s brain. In the nervous system of the worm, neurons communicate with one another through nonlinear, time-varying synaptic connections that are created between them by their unique wiring arrangements. This permits the network to generate complicated behaviours with a minimal number of neurons and allows neurons to exhibit liquid time-constant dynamics. In order to create compact neuronal network topologies to control sequential robotic tasks, we identify neuron-pair communication motifs as design operators. The networks' hierarchical architecture, which extends from sensory neurons through recurrently wired interneurons to motor neurons, is purposefully designed to translate information to action (Lanza et. al, 2020). In this project, we adapt a *C. Elegans* network to derive second-order information about bias statements and generate accurate, generalizable outcomes.

## 3 Proposed Approach

The following steps were taken to construct a neural network to assess the political bias of various news sources.

### 3.1 Converting News Articles & Ratings to Vectors

First, over ten-thousand news articles were scraped from the website *AllSides.com* using python modules BeautifulSoup, urlopen, and newspaper3k over a two-month period. *AllSides* classifies news articles from the internet on a five-point political bias scale using professional teams and community reviews. This project seeks to identify hyper-partisan sources of bias, so only very conservative $(-2)$, neutral $(0)$, or very liberal $(+2)$ articles were used; this step also ensured a larger dataset for more accurate predictions. Accordingly, articles were labeled $-1$ (hyper-partisan liberal), 0 (neutral), or 1 (hyper-partisan conservative). These articles were pre-processed with the following steps:

**Task #1: Text Cleaning** – A customized Python module was created to address sources of distortion in news articles specifically. New lines were replaced with spaces, contractions were expanded, and common stop words (with an expanded custom dictionary commonly in news articles) were removed. Furthermore, HTML tags, advertisements, and special characters were removed from the articles' content.

**Task #2: Text to Sequences** – Using Tinkercad's Universal Sentence Encoder, the article's content was tokenized, and a word dictionary was generated. Each article was converted into a 512×1 vector.

### 3.2 Constructing Networks for Bias Classification

We construct six network seeds: (1) a deep neural network with feedforward connections, (2) a "mesh" network whose parameters are randomly seeded, (3) a subset of the "mesh" network whose network parameters are copied from the first DNN seed and evolved, permitting the DNN to develop non-linear computational dynamics, (4) a subset of the "mesh" network whose synaptic configuration is confined to the *C. Elegans* nematode, (5) a subset of the "mesh" network synaptic configuration begins with the *C. Elegans* nematode's structure but is permitted to evolve beyond the nematode's sparse neural connections, (6) a subset of the "mesh" network which replicates the *C. Elegans* configuration in the fourth network seed which adds the deep neural network from the third network configuration as an input.

1. **Network #1 – Deep Neural Network:** We implement three DNN iterations, varying the number of neurons in the hidden layers. The first iteration had four hidden layers 128, 64, 32, and 16 neurons, the second iteration had four hidden layers with 64, 32, 16, and 10 neurons, and the third iteration had five hidden layers with 256, 128, 64, 32, and 10 neurons. We use the sigmoid function to calculate the output and a generic backpropagation algorithm to train the network.

2. **Network #2 – Randomly Seeded Mesh Neural Network (MNN):** We create a novel class of neural networks that we term "Mesh Neural Network" or "MNN". This model was created to remove the constraint of feed-forward or only recurrent connections. The mesh structure is similar to a Hopfield network, but each node in the "mesh" is recurrent. The MNN integrates a *feedback* mechanism that more closely mirrors the visual cortex and improves the inductive capacity of the network in integrating past examples in its decision calculus, modeled after feedback connections in the brain (Caswell, 2016). The network connections in the MNN are not sequential – a neuron in the 3rd "layer" of a MNN can be connected to a neuron in the 1st "layer".

The mesh structure is a superset of previous networks like DNNs, CNNs, and Boolean configurations. The network can be represented by a simple matrix structure consisting of five matrices: input connections, mesh connections, mesh bias, output connections, and output bias.
The output is calculated as follows:

$$out = g((\boldsymbol{outConnect} + \boldsymbol{outBias})(g((\boldsymbol{meshConnect} + \boldsymbol{meshBias}) * (\boldsymbol{inConnect} \cdot x))))$$

Note that since the mesh has recurrent and feedback elements, there is a delay added to the output in order to avoid convergence issues.

Next, we train the network and configure matrix weights. Previous NLP solutions for bias classification tasks have utilized backpropagation, but the inherent feedback within the mesh architecture makes the algorithm computationally inefficient. We can also imagine a blind search; however, the sheer number of possible values renders this virtually impossible. For example, if we choose 10 possible values for each matrix element, then the total number of cases to evaluate equals $10^{784\times240} + 10^{240\times240} + 10^{240} + 10^{240\times10} + 10^{10}$. Therefore, a better search algorithm is necessary. We use a genetic algorithm to search for a better network. The genetic algorithm is described in detail here:

A. Given $n$ input networks, generate *generation number* – $n$ networks with randomized parameters (being exhibitory or inhibitory), with network parameters determined by $\sim Bernoulli(p1)$, $\sim Bernoulli(p2)$, $\sim Bernoulli(p3)$, $\sim Bernoulli(p4)$, and $\sim Bernoulli(p5)$ where $p1, p2, p3, p4$, and $p5$ are the synaptic polarity of each distribution.
B. Isolate top 10 performing networks ($n_t$) to ensure its weights are unchanged from the corresponding accuracy matrix ($a$) from the previous generation ($n_{g-1}$) and to seed the breeding matrix ($n_b$)
C. Sample breed number of pairs ($m_t$) from $n_t$ ($m_t < max\ [n_t, 2]$) and element-multiply by the breed matrix ($m_b$) and $|1 - m_b|$, respectively such that the matrix outputs a random choice between elements 0 and 1 indicating a mutation or no mutation.
D. Randomly sample of 10 networks is done from the $[0, 90]$ percentile of the connections matrix to ensure sufficient synaptic disparity or ‘population diversity’ for the genetic algorithm.
E. Randomly mutate the number of matrices from the generation by creating mutation matrix ($m_m$) with connections $\sim Bernoulli\ (p_m)$ and returning the sum of that matrix and the original network matrix for each matrix in the network, with a different $m_m$ being generated for each matrix.

Once the “mother” and “father” matrix are chosen at random from the top performing matrices, each element of the “bred” matrix is randomly chosen from the mother and father. This generates the bred elements of the next generation. The images in Figure 1 show an example of the ‘evolution’ of the MNN. Note that the connections themselves are changing along with the weights.

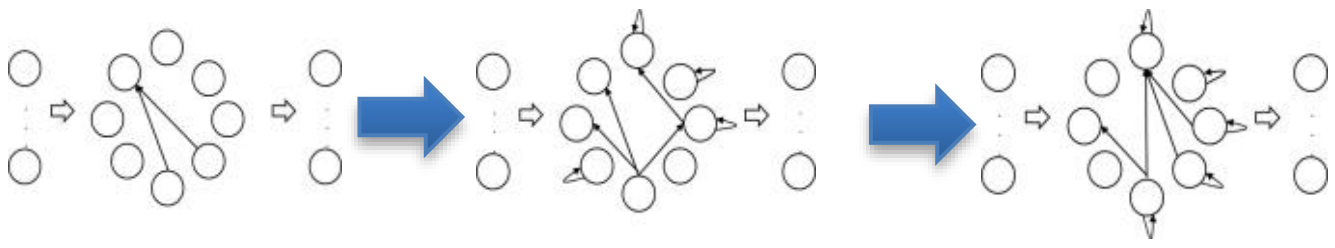

**Figure 1:** Examples of MNN evolution.

For the randomly seeded configuration, we generate a randomly parameterized set of networks to seed the genetic to test the efficacy of the genetic algorithm without a pre-trained structure or network. The randomly-seeded mesh network operates as a control for network performance – the hypothetical gains in performance that it derives would be solely a result of the genetic algorithm. The network’s parameters were initialized using the following Bernoulli distribution:

$$f(x) = \begin{cases} p^x \times (1-p)^{1-x} \\ 0 \end{cases}$$

$$if\ x = 0,1, \begin{cases} 0, & if\ x = 1 \\ 1 - p, & if\ x = 0 \end{cases}$$

**3. Network #3 – Deep Neural Network Seeded in Mesh Neural Network:**

The DNN with initialized parameters from Network #1 is integrated into the MNN structure and evolved with a genetic algorithm. Since the DNN is a subset of the MNN, we can implement this by placing the network weights at specific locations in the mesh connections matrix. A TensorFlow Keras module is used to train the DNN as inputs to the MNN. This is illustrated below:

| | 0 | ... | 127 | 128 | ... | 191 | 192 | ... | 223 | ... | 239 |
|---|---|---|---|---|---|---|---|---|---|---|---|
| **0** | 0 | 0 | 0 | 0 | 0 | 0 | 0 | 0 | 0 | 0 | 0 |
| **...** | 0 | 0 | 0 | 0 | 0 | 0 | 0 | 0 | 0 | 0 | 0 |
| **127** | 0 | 0 | 0 | 0 | 0 | 0 | 0 | 0 | 0 | 0 | 0 |
| **128** | θ1 | θ1 | θ1 | 0 | 0 | 0 | 0 | 0 | 0 | 0 | 0 |
| **129** | θ1 | θ1 | θ1 | 0 | 0 | 0 | 0 | 0 | 0 | 0 | 0 |
| **...** | θ1 | θ1 | θ1 | 0 | 0 | 0 | 0 | 0 | 0 | 0 | 0 |
| **191** | θ1 | θ1 | θ1 | 0 | 0 | 0 | 0 | 0 | 0 | 0 | 0 |
| **192** | 0 | 0 | 0 | θ2 | θ2 | θ2 | 0 | 0 | 0 | 0 | 0 |
| **193** | 0 | 0 | 0 | θ2 | θ2 | θ2 | 0 | 0 | 0 | 0 | 0 |
| **...** | 0 | 0 | 0 | θ2 | θ2 | θ2 | 0 | 0 | 0 | 0 | 0 |
| **222** | 0 | 0 | 0 | θ2 | θ2 | θ2 | 0 | 0 | 0 | 0 | 0 |
| **223** | 0 | 0 | 0 | θ2 | θ2 | θ2 | 0 | 0 | 0 | 0 | 0 |
| **224** | 0 | 0 | 0 | 0 | 0 | 0 | θ3 | θ3 | θ3 | 0 | 0 |
| **...** | 0 | 0 | 0 | 0 | 0 | 0 | θ3 | θ3 | θ3 | 0 | 0 |
| **239** | 0 | 0 | 0 | 0 | 0 | 0 | θ3 | θ3 | θ3 | 0 | 0 |

**Figure 2:** Network weight matrix comprising of 0, 1, $\theta 1, \theta 2\ and\ \theta 3$. The x-coordinate denotes the node that a synapse originates from, and the y-coordinate denotes the node that a synapse connects to.

**4. Network #4 – *C. Elegans* Network:**

We construct a novel, end-to-end recursive time-feedback network structure modelled after *C. Elegans*. Many neural circuits within the nematode’s central nervous systems are constructed by a distinct four-layer hierarchal network topology. They receive environmental or contextual observations from sensory neurons. These are passed on to inter-neurons and command neurons, which generate an output decision. Finally, this decision is passed to the motor neurons to actuate its muscles. The wiring diagram of *C. Elegans* achieves a sparsity of around 90% with predominantly feedforward connections from sensors to intermediate neurons, highly recurrent connections among inter-neurons and command neurons, and feedforward connections from command neurons to motor neurons. This specific topology has attractive computational advantages, such as efficient distributed control, requiring a small number of neurons, hierarchical temporal dynamics, and maximal information propagation in sparse-flow networks. At its core, the network possesses a nonlinear time-varying synaptic transmission mechanism that improves its expressive power and computational efficiency compared with their deep learning counterparts. The architecture of the novel network was determined by the following design principles, illustrated below:

A. Insert four neural layers – $N_s$ sensory neurons, $N_i$ inter-neurons, $N_c$ command neurons and $N_m$ motor neurons.

B. Between every two consecutive layers – ∀ source neuron, insert $n_{so-t}$ synapses ($n_{so-t} \leq N_t$ ), with synaptic polarity ($\sim Bernoulli\ p_2$ ), to $n_{so-t}$ target neurons, randomly selected $\sim Binomial(n_{s(o)-t}, p_1)$ where $n_{so-t}$ is the number of synapses from source to target and $p_1$ and $p_2$ are probabilities corresponding to their distributions.

C. Between every two consecutive layers ∀ target neuron $j$ with no synapse, insert $m_{so-t}$ synapses ($m_{so-t} \leq 1$; $i \neq j$, $L_{ti}$), where $L_{ti}$ is the number of synapses to target neuron $i$, with synaptic polarity (being excitatory or inhibitory) $\sim Bernoulli(p_2)$, from $m_{so-t}$ source neurons, randomly selected from $\sim Binomial(m_{so-t}, p_3)$. $m_{so-t}$ is the number of synapses from source to target neurons with no synaptic connections.

D. In recurrent connections of command neurons – ∀ command neuron, insert $l_{so-t}$ synapses ($l_{so-t} \leq N_c$), with synaptic polarity $\sim Bernoulli(p_2)$, to $l_{so-t}$ target command neurons, randomly selected from $\sim Binomial(l_{so-t}, p_4)$. $l_{s(o)-t}$ is the number of synapses from one interneuron to target neurons.

Notice that the *C. Elegans* network is seeded into the mesh configuration, but the genetic algorithm was confined to the *C. Elegans* parameters – it could not evolve the connections and matrix weights outside of the neural network, so most of the weights in the mesh connections matrix were held constant at 0.

5. **Network #5 – *C. Elegans* Seeded Mesh Network:**

   The *C. Elegans* network from Network #4 was fed into the genetic algorithm, except it wasn't constrained to its local structure. The genetic algorithm was given the capacity to evolve its network beyond the sparse neural capacitors of the *C. Elegans* nematode up to the $240 \times 240$ mesh connections matrix.

6. **Network #6 – *C. Elegans* and DNN Seeded Mesh Network:**

   The *C. Elegans* network from Network #5 and the DNN from Network #2 were both fed as seeds to the genetic algorithm in addition to $Bernoulli$ initialized random networks and evolved.

In conjunction with the genetic algorithm, we test 750 configurations of each mesh matrix, varying: a) the number of mutations in the algorithm $(10, 20, \ldots, 90, 100)$, b) the standard deviation of the mutations matrix $(0.2, 0.4, \ldots, 0.8, 1)$, c) the number of generations $(100, 500, 1000)$, and d) the standard deviation of the randomly generated networks $(0.2, 0.4, \ldots, 0.8, 1)$.

Afterwards, the optimized networks were each fed into a separate "master" genetic algorithm as seeds in order to combine the favourable characteristics of each type of optimized network.

### 3.3 Political Bias Classification on News Sources

Last, we isolate the best performing network configuration to classify the political bias of news articles. Five of the most popular sources of political news in the U.S. were tested: Google News, Fox News, CNN, NPR, and the New York Times. Over a 50-day period, the articles on the first page of the politics section on each news network was scraped and fed as an input into the network to ascertain a political bias classification for that day. Also quantified was search order bias — a phenomenon where some news sources may have an equal content of conservative and liberal articles, but the order in which they are presented may favour one side over the other. This 'bias' was quantified in a study conducted by Robert Epstein et. al (2015); the results of this study were used to normalize article bias readings:

*Given* VCF =
$$\frac{\sum_{i=1}^{n}\begin{cases}1, & A_{i_{pre}}-B_{i_{pre}}>0 \wedge A_{i_{post}}-B_{i_{post}}<0\\ 1, & A_{i_{pre}}-B_{i_{pre}}<0 \wedge A_{i_{post}}-B_{i_{post}}>0\\ 0, & Otherwise\end{cases}}{n}, \text{ and}$$

SEME = Bernoulli (VCF[$S_{i\text{-SEME}(i-1)}$, $S_r$],
Result Normalizing Factor (RNF) =
SEME(VCF[$A_{i_{pre}} = S_{r(n)}$]) × $S_{r\text{-}p}$, where $S$ represents the result classified in position $r$, $A_{i_{pre}}$ represents the preference for A before treatment, $A_{i_{post}}$ represents the preference for A after treatment, $B_{i_{pre}}$ represents the preference for B before treatment, and $B_{i_{post}}$ represents the preference for B after treatment.

## 4 Results and Conclusion

First, we analyze the performance of the DNN trained with gradient descent as a control variable for the genetic algorithm and the *C. Elegans* network. We find that the best variant of the DNN network attains 44.67% accuracy – roughly above the probability of guessing a bias classification.

The following scatter plot illustrates test set accuracy with parameter variations from the three aforementioned network configurations: C. Elegans, C. Elegans Seeded Mesh Network, and C. Elegans and DNN Seeded Mesh Network.

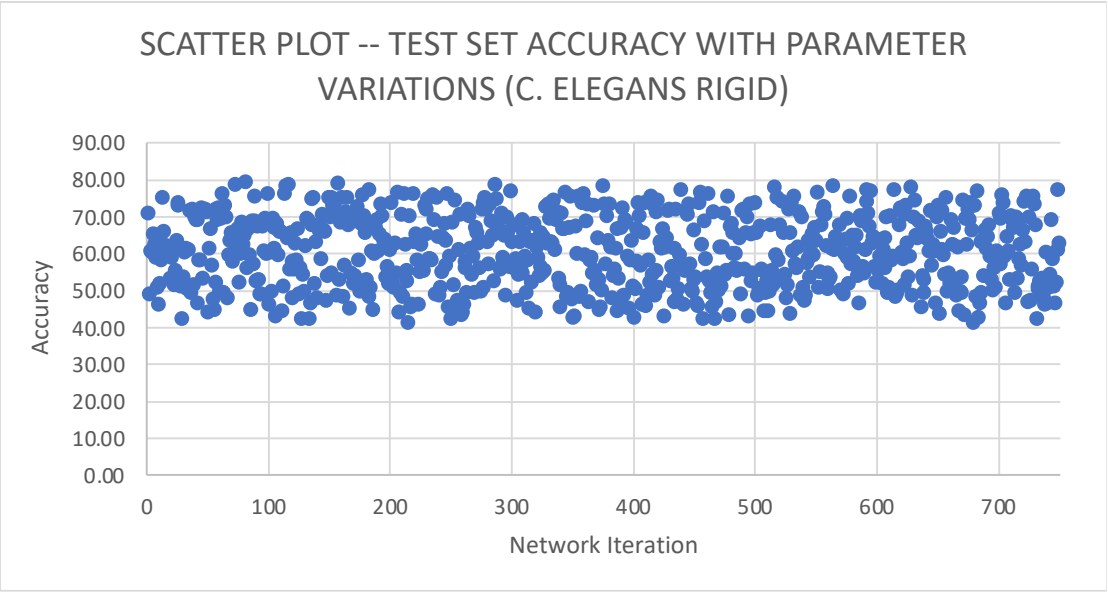

**Figure 3:** Plot of network iteration vs. accuracy for *C. Elegans* Rigid. The plot compares network performance of Network 3 (*C. Elegans* Network) with the number of iterations the genetic algorithm was run for.

**Figure 4:** Plot of Network Iteration Vs. Accuracy for *C. Elegans* Seed. The plot compares network performance of Network 3 (*C. Elegans* Seeded Mesh Network) with the number of iterations the genetic algorithm was run.

**Figure 5:** Plot of Network Iteration Vs. Accuracy for *C.*

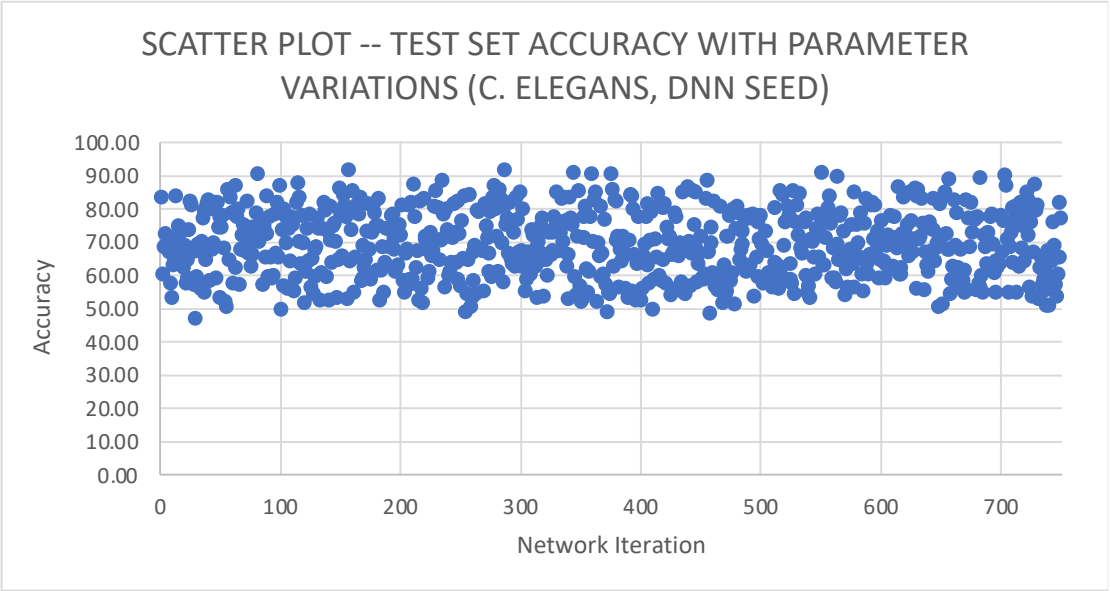

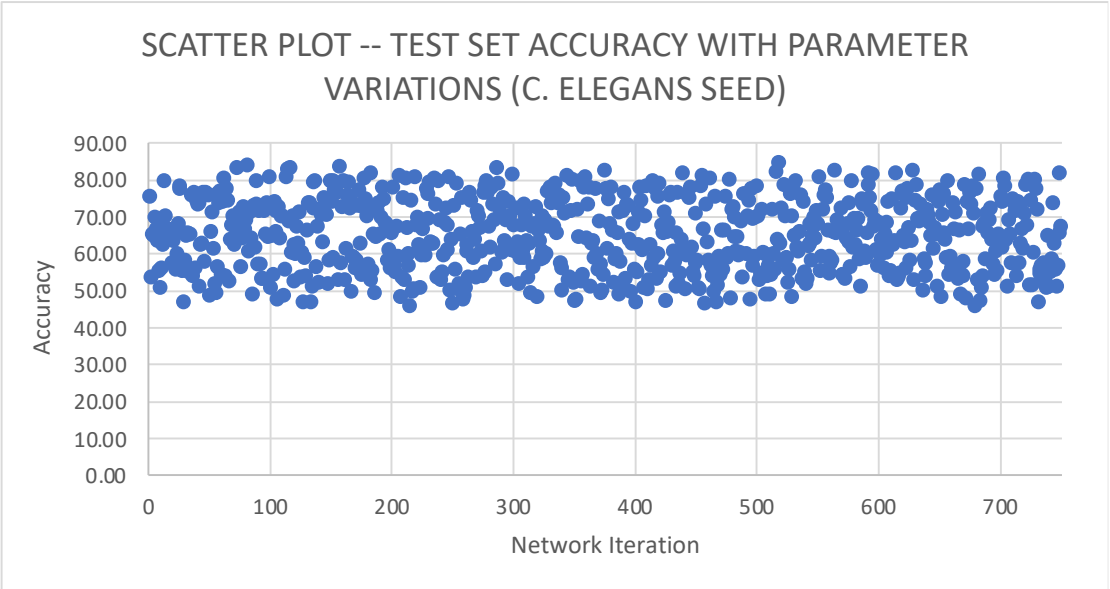

*Elegans* with DNN Seed. The plot compares network performance of Network 3 (*C. Elegans* and DNN Seeded Mesh Network) with the number of iterations the genetic algorithm was run.

Interestingly, we see that the best-performing networks are all variants of the *C. Elegans* class – this suggests that the incorporation of the nematode into the network decision calculus provides increasingly generalizable results. Thus, the *C. Elegans* network structure that has been evolved through natural processes for centuries trumps contemporary artificial network structures. Furthermore, the C. elegans performed ~5% better when it was used as a seed to the MNN superstructure, and the C. elegans combined with the trained DNN performed ~8% better than the *C. Elegans* seed. This suggests that although the *C. Elegans* structure can reasonably operate in difficult classification tasks, it may be more effective when it is: 1) Given the capacity to evolve its own network structure 2) Evolve in conjunction with other networks, to obtain more genetic variation.

We observe an incredible amount of variation in terms of network accuracy when parameters are varied in the graph. Recall that the parameters used for this project are number of mutations, standard deviation of mutations, number of generations, and standard deviation of random nets. Therefore, we use the Pearson correlation coefficient to quantify the correlation between each factor and network performance. The following table displays r-coefficients for each network is displayed below:

| Network Type | Mutations #/Test Accuracy Correlation | Std. Dev./Test Accuracy Correlation | Generation #/Test Accuracy Correlation | Random Std. Dev/Test Accuracy Correlation |
|---|---|---|---|---|
| Randomly Seeded | -0.01301 | 0.02205 | -0.05582 | 0.045212 |
| *C. Elegans Rigid* | -0.04236 | -0.00383 | 0.015519 | -0.00129 |
| *C. Elegans* Seeded Mesh Network | -0.04195 | -0.00078 | 0.017185 | 0.000774 |
| *C. Elegans and DNN* Seeded Mesh Network | -0.03337 | -0.01167 | 0.010474 | -0.00292 |

**Table 1:** Correlation between network parameters and performance.

Surprisingly, we see that none of the factors have a strong correlation with network performance, as each r value is in the $[-0.5, 0.5]$ range. This may be attributed to the genetic algorithm and mutation functions causing increased variation, which cancels out the impact of these factors. Regardless, we find that the best sub-parameters were: 40 mutations, 1 Std. Dev., 100 Generations, 0.2 Std. Dev. (Random Nets). This configuration returned ~92% test set accuracy. We then use these network parameters to quantify accuracy percentage by generation with all three network configurations, shown below:

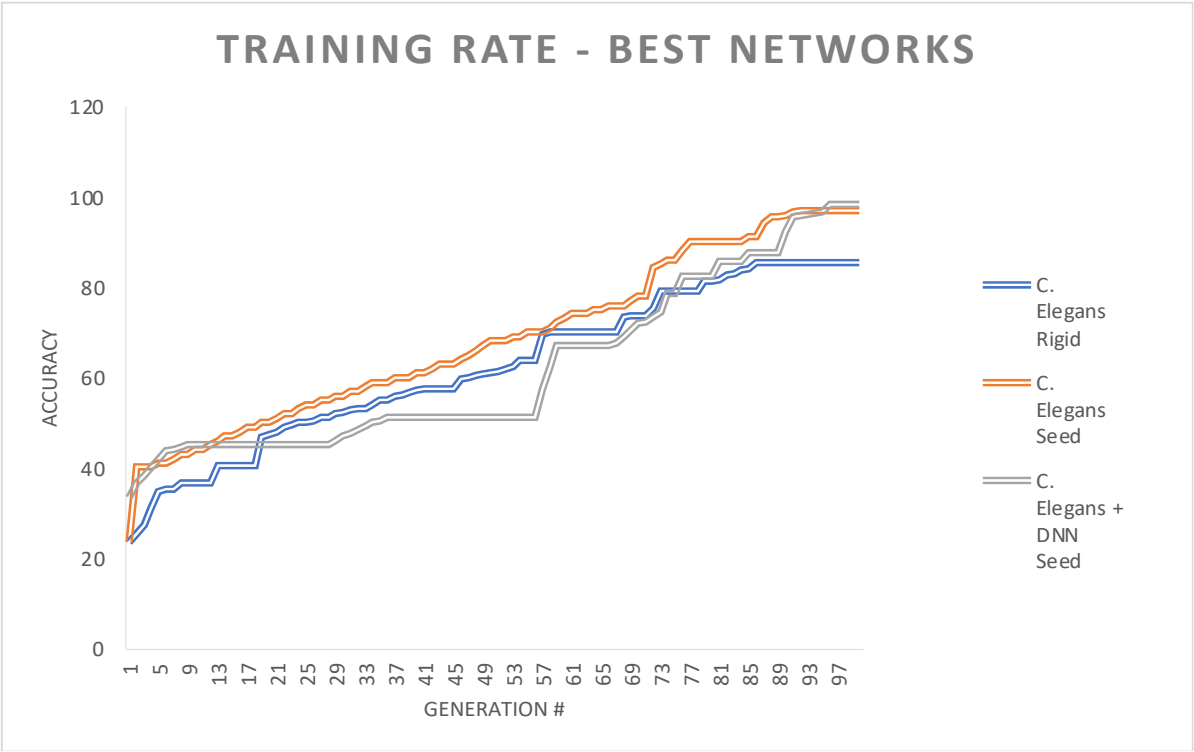

**Figure 6:** Plot of generations vs. accuracy for training rates of best networks.

Interestingly, we see that the training rate follows a "stair-step" pattern – there is a sharp increase, and then the network accuracy stagnates, which is then followed by a sharp increase. The primary difference between the *C. Elegans* networks is the point and rate at which at which the increases occur. The worst performing network – *C. Elegans* rigid – does not improve in the last twenty generations – meaning the genetic algorithm did not find a favorable matrix upon multiple evolutions. This mirrors Darwinian evolution – although natural selection is effective – its ability to produce more resilient species is constrained by initial biological variation. In this case, the constraint is that the genetic algorithm is limited to the *C. Elegans* structure, which offers a finite number of possibilities for mutation. The next network – the *C. Elegans* Seed – performs much better – it does not significantly stagnate in the latter generations. This reflects the network structure – when the genetic algorithm can mutate elements within the 240×240 structure rather than the isolated *C. Elegans*, performance increases. This proves that the nematode is a favorable starting point for the network, but other structures are more effective in the process of training with the genetic algorithm. Lastly, the *C. Elegans* & DNN Seed performs the best. The same trend is evident here – the addition of the DNN allows the genetic algorithm to incorporate more favorable elements from the "parents" to the "children" – preventing stagnation and allowing for more effective mutations.

Next, we ascertain if our findings are statistically significant. The Shapiro-Wilk test reveals that the data is not distributed normally, although the majority of the data points are mesokurtic. Therefore, we utilize the Wilcoxon Signed-Rank test for statistical significance – a non-parametric version of the paired t-test. The table below isolates the results of the test, in which the Random network was used as the control. Comparing performance vs. the random network isolates the performance of the networks as it relates to the genetic algorithm, and to some degree 'cancels out' the impact of the genetic algorithm on network performance.

| | |
|---|---|
| *C. Elegans*—Rigid | 0.000000165 |
| *C. Elegans*—Seed | Less than 0.000000001 |
| *C. Elegans* & DNN—Seed | Less than 0.000000001 |

**Table 2**: Network Statistical Significance Tests

Thus, we can decisively state that the change in network performance was statistically significant, and the addition of the C. elegans, MNN, and genetic algorithm substantially increases network accuracy.

Last, we utilize the best-performing network to ascertain the bias of the following news sources: Google News, Fox News, CNN, NPR, and the New York Times. Table 3 shows the average bias classifications obtained over the 50-day trial period for Google News after adjusting for search order bias.

| Google News | Fox News | CNN | NPR | New York Times |
|---|---|---|---|---|
| -0.1618 | 0.8934 | -0.7939 | -0.2132 | -0.4623 |

**Table 3**: Average Bias Classifications

There are three central conclusions that our findings reveal:

First, the addition of a recursive feedback mechanism in the novel network structure – MNN – outperforms state of the art networks and improves generalizability by an exceptionally large margin. Previous studies using DNN and CNN networks and the DNN trained in this study did not surpass the 50% threshold. Utilizing a genetic algorithm to train the MNN and the C. elegans matrix drastically improves network performance. The best MNN network – the MNN seeded with a DNN and C. elegans – improved to 92% test set accuracy – roughly human-level performance!

Second, the initial seed of the genetic algorithm has a substantial effect on network performance – the Random control network had an accuracy percentage of 53.9%, whereas the C. elegans and DNN Seed network had an accuracy percentage of 92% – this is similar to how there must be favorable genotypical traits in a population to produce efficient species through natural selection.

Third, of the five most popular news sources, only three could be considered somewhat neutrally biased. We can define above 0.5 or below 0.5 to indicate significant bias in news sources, and two news sources are above that parameter: CNN and Fox News.